\documentclass[11pt]{article}
\usepackage{acl2015}
\usepackage{times}
\usepackage{url}
\usepackage{latexsym}
\usepackage{graphicx}
\usepackage{sidecap}
\usepackage{amsfonts}
\usepackage[export]{adjustbox}
\usepackage{makecell}
\usepackage{multirow}
\usepackage{url}
\usepackage{amsmath}
\usepackage{sidecap}
\usepackage{multirow}
\usepackage[export]{adjustbox}
\usepackage{times}
\usepackage{url}
\usepackage{latexsym}
\usepackage{graphicx}
\usepackage{amsmath}
\usepackage{mathtools}
\usepackage{color}
\usepackage{multirow}
\usepackage{algorithmicx}
\usepackage{algorithm}
\usepackage{algpascal}
\usepackage{algc}
\usepackage{algcompatible}
\usepackage{algpseudocode}
\usepackage{breqn}
\usepackage{float}
\usepackage{anyfontsize}
\usepackage{verbatim}

\DeclareMathOperator*{\argmax}{arg\,max}

\newcommand{\pathSeparator}{$\rightarrow$ }

\title{Compositional Vector Space Models for Knowledge Base Completion}
\author{Arvind Neelakantan, Benjamin Roth, Andrew McCallum  \\
  Department of Computer Science \\
University of Massachusetts, Amherst \\ Amherst, MA, 01003 \\
  {\tt \{arvind,beroth,mccallum\}@cs.umass.edu} \\
 }
\date{}

\begin{document}
\maketitle
\begin{abstract}
Knowledge base (KB) completion adds new facts to a KB by making
inferences from existing facts, for example by inferring with high
likelihood {\it nationality(X,Y)} from {\it bornIn(X,Y)}.  Most
previous methods infer simple one-hop relational synonyms like this,
or use as evidence a multi-hop relational path treated as an atomic
feature, like {\it bornIn(X,Z) \pathSeparator containedIn(Z,Y)}.  This
paper presents an approach that reasons about conjunctions of
multi-hop relations {\sl non-atomically}, composing the implications
of a path using a recurrent neural network (RNN) that takes as inputs
vector embeddings of the binary relation in the path.  Not only does
this allow us to generalize to paths unseen at training time, but
also, with a single high-capacity RNN, to predict new relation types
not seen when the compositional model was trained (zero-shot
learning).  We assemble a new dataset of over 52M relational triples,
and show that our method improves over a traditional classifier by
11\%, and a method leveraging pre-trained embeddings by 7\%.
\end{abstract}

\section{Introduction}

Constructing large knowledge bases (KBs) supports downstream reasoning
about resolved entities and their relations, rather than the noisy
textual evidence surrounding their natural language mentions.  For
this reason KBs have been of increasing interest in both industry and
academia \cite{freebase,yago,NELL}.  Such KBs typically contain many
millions of facts, most of them (entity1,relation,entity2) ``triples''
(also known as binary relations) such as \emph{(Barack Obama,
  presidentOf, USA)} and \emph{(Brad Pitt, marriedTo, Angelina
  Jolie)}.

However, even the largest KBs are woefully incomplete
\cite{min2013distant}, missing many important facts, and therefore
damaging their usefulness in downstream tasks.  Ironically, these
missing facts can frequently be inferred from other facts already in
the KB, thus representing a sort of inconsistency that can be repaired
by the application of an automated process.  The addition of new
triples by leveraging existing triples is typically known as {\it KB
  completion}.

Early work on this problem focused on learning symbolic rules.  For
example, \newcite{horn_clauses} learns Horn clauses predictive of new
binary relations by exhausitively exploring relational paths of
increasing length, and selecting those surpassing an accuracy
threshold.  (A ``path'' is a sequence of triples in which the second
entity of each triple matches the first entity of the next triple.)
\newcite{pra} introduced the Path Ranking Algorithm (PRA), which
greatly improves efficiency and robustness by replacing exhaustive
search with random walks, and using unique paths as features in a
per-target-relation binary classifier.  A typical predictive feature
learned by PRA 
is that \emph{CountryOfHeadquarters(X, Y)} is implied by
\emph{IsBasedIn(X,A)} and \emph{StateLocatedIn(A, B)} and
\emph{CountryLocatedIn(B, Y)}.  Given \emph{IsBasedIn(Microsoft,
  Seattle)}, \emph{StateLocatedIn(Seattle, Washington)} and
\emph{CountryLocatedIn(Washington, USA)}, we can infer the fact
\emph{CountryOfHeadquarters(Microsoft, USA)} using the predictive feature.  In
later work, \newcite{pra_second} greatly increase available raw
material for paths by augmenting KB-schema relations with relations defined by
the text connecting mentions of entities in a large corpus 
(also known as OpenIE relations \cite{openie}).

However, these symbolic methods can produce many millions of distinct
paths, each of which is categorically distinct, treated by PRA as a
distinct feature.  (See Figure~1.)  Even putting aside the OpenIE
relations, this limits the applicability of these methods to modern
KBs that have thousands of relation types, since the number of
distinct paths increases rapidly with the number of relation types.
If textually-defined OpenIE relations are included, the problem is
obviously far more severe.

Better generalization can be gained by operating on embedded vector
representations of relations, in which vector
similarity can be interpreted as semantic similarity.  For example,
\newcite{transe} learn low-dimensional vector representations of
entities and KB relations, such that vector differences between two
entities should be close to the vectors associated with their
relations.  This approach can find relation synonyms, and thus perform
a kind of one-to-one, non-path-based relation prediction for KB
completion.  Similarly \newcite{rescal} and \newcite{socherkb} perform
KB completion by learning embeddings of relations, but based on
matrices or tensors.  Universal schema \cite{limin} learns to
perform relation prediction cast as matrix completion (likewise using
vector embeddings), but predicts textually-defined OpenIE relations as
well as KB relations, and embeds entity-pairs in addition to
individual entities.  Like all of the above, it also reasons about
individual relations, not the evidence of a connected path of relations.

This paper proposes an approach combining the advantages of (a)
reasoning about conjunctions of relations connected in a path, and (b)
generalization through vector embeddings, and (c) reasoning non-atomically
and compositionally about the elements of the path, for further
generalization.

Our method uses recurrent neural networks (RNNs) \cite{recurrentnn}
to compose the semantics of relations in an arbitrary-length path.  At
each path-step it consumes both the vector embedding of the next
relation, and the vector representing the path-so-far, then outputs a
composed vector (representing the extended path-so-far), which will be
the input to the next step.  After consuming a path, the RNN should
output a vector in the semantic neighborhood of the relation between the
first and last entity of the path.  For example, after consuming the
relation vectors along the path \emph{Melinda Gates \pathSeparator Bill Gates \pathSeparator
  Microsoft \pathSeparator Seattle}, our method produces a vector very close to the
relation \emph{livesIn}.


Our compositional approach allow us at test time to make predictions
from paths that were unseen during training, because of the
generalization provided by vector neighborhoods, and because they are
composed in non-atomic fashion.  This allows our model to seamlessly
perform inference on many millions of paths in the KB graph.  In most
of our experiments, we learn a separate RNN for predicting each
relation type, but alternatively, by learning a single high-capacity
composition function for all relation types, our method can perform
zero-shot learning---predicting new relation types for which the
composition function was never explicitly trained.

\begin{figure}[t]
\centering
\includegraphics[scale=0.25]{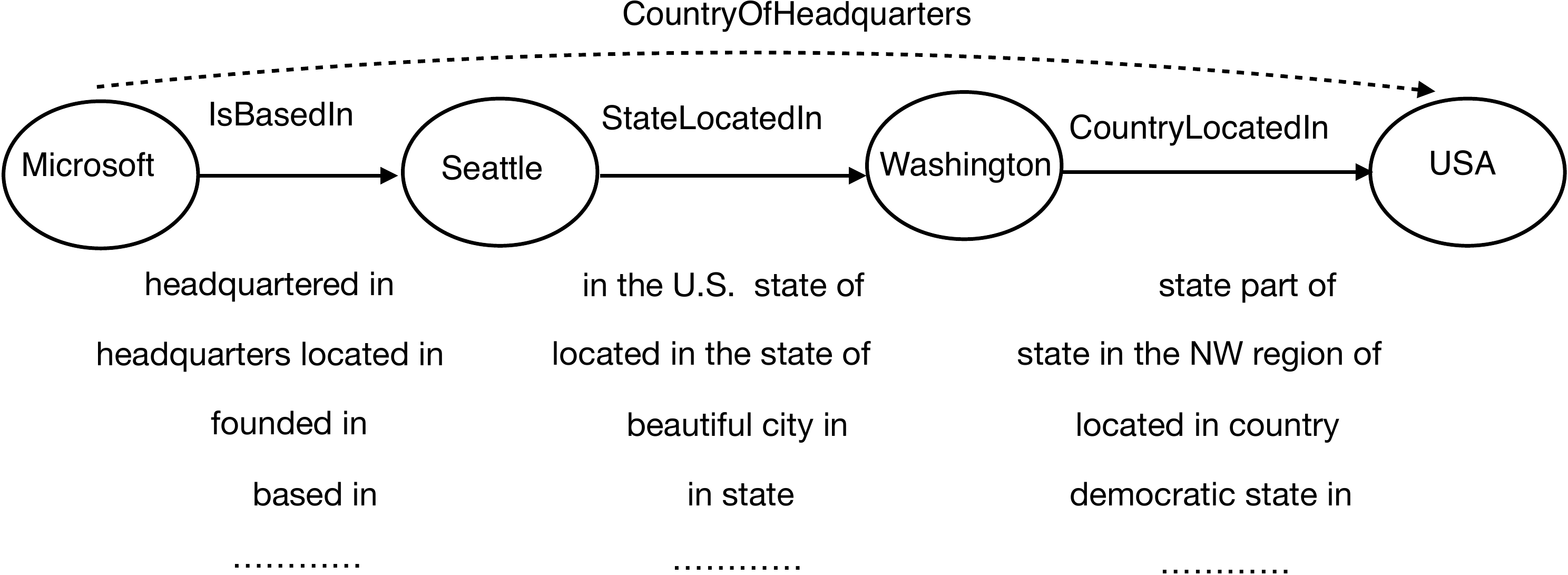}
\caption{\small Semantically similar paths connecting entity pair (Microsoft, USA).}
\label{example}
\end{figure}

Related to our work, new versions of PRA \cite{pra_recent,vector_pra}
use pre-trained vector representations of relations to alleviate
its feature explosion problem---but the core mechanism continues to be
a classifier based on atomic-path features.  In the 2013 work many
paths are collapsed by clustering paths according to their relations'
embeddings, and substituting cluster ids for the original relation
types.  In the 2014 work unseen paths are mapped to nearby paths seen
at training time, where nearness is measured using the embeddings.
Neither is able to perform zero-shot learning since there must be a
classifer for each predicted relation type.  Furthermore their
pre-trained vectors do not have the opportunity to be tuned to the KB
completion task because the two sub-tasks are completely disentangled.

An additional contribution of our work is a new large-scale data set
of over 52 million triples, and its preprocessing for purposes of
path-based KB completion (can be downloaded from \url{ http://iesl.cs.umass.edu/downloads/inferencerules/release.tar.gz}).
The dataset is build from the combination of Freebase \cite{freebase}
and Google's entity linking in ClueWeb \cite{clueweb}.  Rather than
Gardner's 1000 distinct paths per relation type, we have over 2
million.  Rather than Gardner's 200 entity pairs, we use over 10k.
All experimental comparisons below are performed on this new data set.


On this challenging large-scale dataset our compositional method
outperforms PRA \cite{pra_second}, and Cluster PRA \cite{pra_recent}
by 11\% and 7\% respectively.  A further contribution of our work is a
new, surprisingly strong baseline method using classifiers of path
bigram features, which beats PRA and Cluster PRA, and statistically
ties our compositional method.  Our analysis shows that our method has
substantially different strengths than the new baseline, and the
combination of the two yields a 15\% improvement over
\newcite{pra_recent}.  We also show that our zero-shot model is indeed
capable of predicting new unseen relation types.

\begin{figure}[t]
\centering
\includegraphics[scale=0.17]{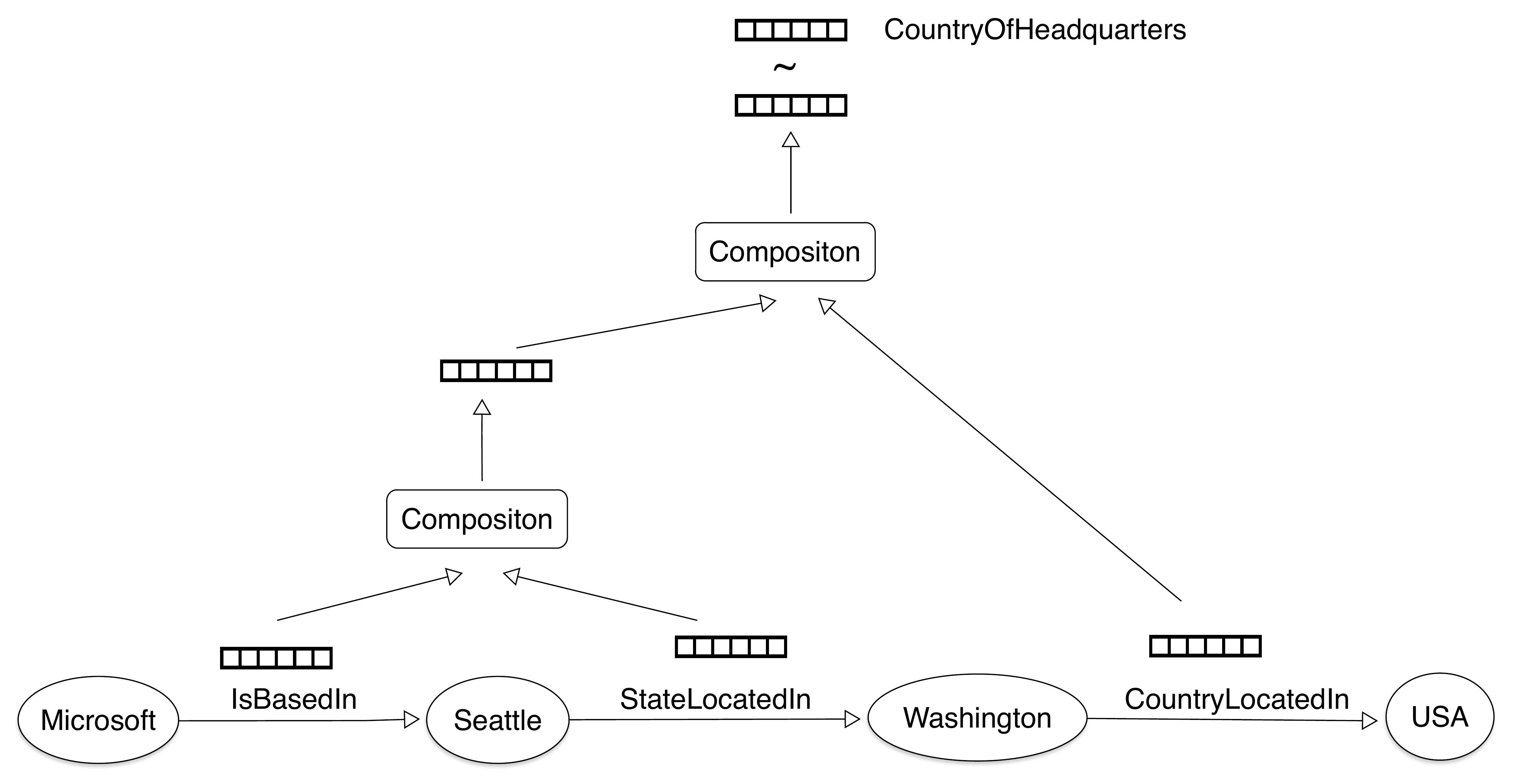}
\caption{\small Vector Representations of the paths are computed by applying the composition function recursively.}
\label{rnn}
\end{figure}

\section{Background}
We give background on PRA which we use to obtain a set of paths connecting the entity pairs and the RNN model  which we employ to model the composition function. 

\subsection{Path Ranking Algorithm}
Since it is impractical to exhaustively obtain the set of all paths connecting an entity pair in the large KB graph, we use PRA \cite{pra} to obtain a set of paths connecting the entity pairs.  Given a training set of entity pairs for a relation, PRA heuristically finds a set of paths by performing random walks from the source and target nodes keeping the most common paths.
We use PRA to find millions of distinct paths per relation type. We do not use the random walk probabilities given by PRA since using it did not yield improvements in our experiments.  

\subsection{Recurrent Neural Networks}
Recurrent neural network (RNN) \cite{recurrentnn} is a neural network that constructs vector representation for sequences (of any length). For example, a RNN model can be used to construct vector representations for phrases or sentences (of any length) in natural language by applying a composition function \cite{rnnlm,rnnmt,rnnparse}. The vector representation of a phrase ($w_{1}$, $w_{2}$)  consisting of  words $w_{1}$ and $w_{2}$ is given by $f(W [v(w_{1}) ; v(w_{2}) ])$ where $v(w) \in \mathbb{R}^{d}$  is the vector representation of $w$, $f$ is an element-wise non linearity function, $[a ; b]$ represents the concatenation two vectors $a$ and $b$ along with a bias term,  and $W \in  \mathbb{R}^{d \times 2*d+1}$ is the composition matrix. This operation can be repeated to construct vector representations of longer phrases.

\section{Recurrent Neural Networks for KB Completion}
\label{sec:rnn}
This paper proposes a RNN model for KB completion that reasons on the paths connecting an entity pair to predict missing relation types. The vector representations of the paths (of any length) in the KB graph are computed by applying the composition function recursively as shown in Figure \ref{rnn}. To compute the vector representations for the higher nodes in the tree, the composition function consumes the vector representation of the node's two children nodes and outputs a new vector of the same dimension. Predictions about missing relation types are made by comparing the vector representation of the path with the vector representation of the relation using the sigmoid function. 

We represent each binary relation using a $d$-dimensional real valued vector. We model composition using recurrent neural networks \cite{recurrentnn}.  We learn a separate composition matrix for every relation that is predicted.

Let $v_{r}(\delta) \in \mathbb{R}^{d}$ be the vector representation of relation $\delta$ and $v_{p}(\pi) \in \mathbb{R}^{d}$ be the vector representation of path $\pi$. $v_p(\pi)$ denotes the relation vector if path $\pi$ is of length one. To predict relation $\delta=$ \emph{CountryOfHeadquarters}, the vector representation of the path  $\pi=$ \emph{IsBasedIn \pathSeparator StateLocatedIn} containing two relations  \emph{IsBasedIn} and \emph{StateLocatedIn} is computed by (Figure \ref{rnn}),
\begin{align*}
	v_{p}(\pi) &=\\& f(W_{\delta} [v_{r}(\emph{IsBasedIn}) ; v_{r}(\emph{StateLocatedIn}) ]) 
\end{align*}
where $f=sigmoid$ is the element-wise non-linearity function, $W_{\delta} \in \mathbb{R}^{d*2d+1}$ is the composition matrix for $\delta=$ \emph{CountryOfHeadquarters} and 
$[a;b]$ represents the concatenation of two vectors $a \in \mathbb{R}^{d}, b \in \mathbb{R}^{d}$ along with a bias feature to get a new vector $[a;b] \in \mathbb{R}^{2d+1}$. 

The vector representation of the path $\Pi=$ \emph{IsBasedIn \pathSeparator StateLocatedIn \pathSeparator CountryLocatedIn} in Figure \ref{rnn} is computed similarly by,
\begin{align*}
	v_{p}(\Pi) &=\\& f(W_{\delta} [v_{p}(\pi) ; v_{r}(\emph{ CountryLocatedIn}) ]) 
\end{align*}
where $v_{p}(\pi)$ is the vector representation of path \emph{IsBasedIn \pathSeparator StateLocatedIn}. While computing the vector representation of a path we always traverse left to right, composing the relation vector in the right with the accumulated path vector in the left\footnote{we did not get significant improvements when we tried more sophisticated ordering schemes for computing the path representations.}. This makes our model a recurrent neural network \cite{recurrentnn}.

Finally, we make a prediction regarding \emph{CountryOfHeadquarters(Microsoft, USA)}  using the path $\Pi=$ \emph{IsBasedIn \pathSeparator StateLocatedIn \pathSeparator CountryLocatedIn} by comparing the vector representation of the path ($v_{p}(\Pi)$) with the vector representation of the relation $\emph{CountryOfHeadquarters}$ ($v_{r}$(\emph{CountryOfHeadquarters})) using the sigmoid function.  
\newfloat{algorithm}{t}{lop}
\begin{algorithm}
\caption{\small Training Algorithm of RNN model for relation $\delta$}
\begin{algorithmic}[1]
 \State Input: $\Lambda_{\delta} = \Lambda_{\delta}^{+} \cup \Lambda_{\delta}^{-}, \Phi_{\delta}$, number of iterations $T$, mini-batch size $B$
\State Initialize $v_{r}, W_{\delta}$ randomly
 \For{$t = 1, 2, \ldots, T$}
 	\State $\nabla {v_r} = 0$,  $\nabla {W_{\delta}} = 0$ and $b = 0$
	\For{$\lambda =(\gamma, \delta) \in \Lambda_{\delta}$ }
 		\State $\mu_{\lambda} = \argmax_{\pi \in \Phi_{\delta}(\gamma)} v_{p}(\pi).v_{r}(\delta)$
 		\State Accumulate gradients to $\nabla {v_r}$,  $\nabla {W_{\delta}}$ \\  	\hskip\algorithmicindent  using path $\mu_{\lambda}$.
 		 \State $b = b + 1$
 		 \If {$b=B$}
 		 	\State Gradient Update for $v_r, W_{\delta}$
 		 	\State $\nabla {v_r} = 0$,  $\nabla {W_{\delta}} = 0$ and $b = 0$
 		 \EndIf	
 	\EndFor
 	 \If {$b > 0$}
 		 	\State Gradient Update for $v_r, W_{\delta}$
 	\EndIf	
 \EndFor
\State Output: $v_{r}, W_{\delta}$ 
\end{algorithmic}
\end{algorithm}
\subsection{Model Training}
We train the model  with the existing facts in a KB using them as positive examples and negative examples are obtained by treating the unobserved instances as negative examples \cite{distant_supervision,pra,limin,transe}. Unlike in previous work that use RNNs\cite{parsing,qa,irsoy}, a challenge with using them for our task  is that among the set of paths connecting an entity pair, we do not observe which of  the path(s) is predictive of a relation.  We select the path that is closest to the relation type to be predicted in the vector space. This not only allows for faster training (compared to marginalization) but also gives improved performance. This technique has been successfully  used in models other than RNNs previously \cite{weston,mssg}. 

We assume that we are given a KB (for example, Freebase enriched with SVO triples) containing a set of entity pairs $\Gamma$, set of relations $\Delta$ and a set of observed facts $\Lambda^{+}$ where  $\forall \lambda = (\gamma, \delta) \in  \Lambda^{+} (\gamma \in \Gamma, \delta \in \Delta)$ indicates a positive fact that entity pair $\gamma$ is in relation $\delta$. Let $\Phi_{\delta}(\gamma)$ denote the set of paths connecting entity pair $\gamma$ given by PRA for predicting relation $\delta$. 
    
In our task, we only observe the set of paths connecting an entity pair but we do not observe which of the path(s) is predictive of the fact. We treat this as a latent variable ($\mu_{\lambda}$ for the fact $\lambda$)  and we assign $\mu_{\lambda}$ the path whose vector representation has maximum dot product with the vector representation of the relation to be predicted. For example,  $\mu_{\lambda}$ for the fact $\lambda =(\gamma, \delta)  \in  \Lambda^{+} $ is given by, 
\begin{align*}
 \mu_{\lambda} = \argmax_{\pi \in \Phi_{\delta}(\gamma)} v_{p}(\pi).v_{r}(\delta)
\end{align*}    
During training, we assign $\mu_{\lambda}$ using the current parameter estimates. We use the same procedure to assign $\mu_{\lambda}$ for unobserved facts that are used as negative examples during training.  

We train a separate RNN model for predicting each relation and the parameters of the model for predicting relation $\delta \in \Delta$ are $\Theta=\{v_r(\omega) \forall \omega \in \Delta, \; W_{\delta}\}$. Given a training set consisting of positive ($\Lambda_{\delta}^{+}$) and negative ($\Lambda_{\delta}^{-}$) instances\footnote{we sub-sample a portion of the set of all unobserved instances.}   for relation  $\delta$, the parameters are trained to maximize the log likelihood of the training set with L-2 regularization.
\begin{align*}
\Theta^{*} = \argmax_{\Theta} & \sum \limits_{\lambda  =(\gamma, \delta) \in \Lambda_{\delta}^{+}}   P( y_{\lambda} = 1 ; \Theta)  + \\ & \sum \limits_{\lambda  =(\gamma, \delta) \in \Lambda_{\delta}^{-}}   P( y_{\lambda} = 0 ; \Theta)  -  \rho \|  \Theta  \|^2	
\end{align*}    
where $y_{\lambda}$ is a binary random variable which takes the value $1$ if the fact $\lambda$ is true and $0$ otherwise, and the probability of a fact $P( y_{\lambda} = 1 ; \Theta)$ is given by, 
\begin{align*}
P( y_{\lambda} = 1 ; \Theta) = sigmoid(v_p(\mu_{\lambda}).v_r(\delta)) \\
where \; \;  \mu_{\lambda} = \argmax_{\pi \in \Phi_{\delta}(\gamma)} v_{p}(\pi).v_{r}(\delta)  
\end{align*}
and $P( y_{\lambda} = 0 ; \Theta)  = 1 - P( y_{\lambda} = 1 ; \Theta)$.
The relation vectors and the composition matrix are initialized randomly. We train the network using backpropagation through structure \cite{backprop}.

\section{Zero-shot KB Completion}
\label{sec:zeroshot}  
The KB completion task involves predicting facts on thousands of relations types and it is highly desirable that a method can infer facts about relation types without directly training for them.   Given the vector representation of the relations, we show that our model described in the previous section is capable of predicting relational facts without explicitly training for the target (or test) relation types (zero-shot learning).

In zero-shot or zero-data  learning \cite{zeroshot,zero},  some labels or classes are not available  during training the model and only a description of those classes are given at prediction time. 
We make two modifications to the model described in the previous section, (1) learn a general composition matrix, and (2) fix relation vectors with pre-trained vectors, so that we can predict relations that are unseen during training. This ability of the model to generalize to unseen relations is beyond the capabilities  of all previous methods for KB inference \cite{horn_clauses,pra,pra_recent,vector_pra}.

We learn a general composition matrix for all relations instead of learning a separate composition matrix for every relation to be predicted.  So, for example,  the vector representation of the path  $\pi=$ \emph{IsBasedIn \pathSeparator StateLocatedIn} containing two relations  \emph{IsBasedIn} and \emph{StateLocatedIn} is computed by (Figure \ref{rnn}),
\begin{align*}
	v_{p}(\pi) &=\\& f(W [v_{r}(\emph{IsBasedIn}) ; v_{r}(\emph{StateLocatedIn}) ]) 
\end{align*}
where $W \in \mathbb{R}^{d*2d+1}$ is the general composition matrix.


We initialize the vector representations of the binary relations ($v_r$) using the representations learned in \newcite{limin} and do not update them during training. The relation vectors are not updated because at prediction time we would be predicting relation types which are never seen during training and hence their vectors would never get updated.  We learn only the general composition matrix in this model. We train a single model for a set of relation types by replacing the sigmoid function with a softmax function while computing probabilities  and the parameters of the composition matrix are learned using the available training data containing instances of few relations. The other aspects of the model remain unchanged.

To predict facts whose relation types are unseen during training, we compute the vector representation of the path using the general composition matrix and compute the probability of the fact using the  pre-trained relation vector. For example, using the vector representation of the path $\Pi=$ \emph{IsBasedIn \pathSeparator StateLocatedIn \pathSeparator CountryLocatedIn}  in Figure \ref{rnn}, we can predict any relation irrespective of whether they are seen at training by comparing it with the pre-trained relation vectors. 

\section{Experiments}
The hyperparameters of all the models were tuned on the same held-out development data.  All the neural network models are trained for $150$ iterations using $50$ dimensional relation vectors, and we set the L2-regularizer and learning rate to $0.0001$ and $0.1$ respectively. We halved the learning rate after every $60$ iterations and use mini-batches of size $20$. The neural networks and the classifiers were optimized using AdaGrad \cite{adagrad}. 
\begin{table}[t!]
{
\centering
\begin{tabular}{ |l|l| }
\hline
Entities & 18M   \\ \hline
Freebase triples & 40M   \\ \hline
ClueWeb triples & 12M   \\ \hline
Relations & 25,994 \\ \hline
Relation types tested & 46  \\ \hline
Avg. paths/relation & 2.3M  \\ \hline
Avg. training facts/relation & 6638   \\ \hline
Avg. positive test instances/relation & 3492   \\ \hline
Avg. negative test instances/relation & 43,160   \\ \hline
\end{tabular}
\caption{Statistics of our dataset.}
\label{table:stat}
}
\end{table}
\subsection{Data}
We ran experiments on Freebase \cite{freebase} enriched with information from ClueWeb. We use the publicly available entity links to Freebase in the ClueWeb dataset  \cite{clueweb}.  Hence, we create nodes only for Freebase entities in our KB graph. We remove facts containing /type/object/type as they do not give useful predictive information for our task.  We get triples from ClueWeb by considering sentences that contain two entities linked to Freebase. We extract  the phrase between the two entities and treat them as the relation types. For phrases that are of length greater than four we keep only the first and last two words. This helps us to avoid the time consuming step of dependency parsing the sentence to get the relation type. These triples are similar to facts obtained by OpenIE \cite{openie}. To reduce noise, we select relation types that occur at least $50$ times. We evaluate on $46$ relation types in Freebase that have the most number of instances. The methods are evaluated on a subset of facts in Freebase that were hidden during training. Table \ref{table:stat} shows important statistics of our dataset.

\subsection{Predictive Paths}
Table \ref{table:paths} shows predictive paths for $4$ relations learned by the RNN model. The high quality of unseen paths is indicative of the fact that the RNN model is able to generalize to paths that are never seen during training.
\begin{table*}
\centering
{\fontsize{8.3pt}{9.3} \selectfont
\begin{tabular}{|l|}
\hline
\textbf{Relation:} \hspace{.2cm} /book/written\_work/original\_language/  \hspace{.2cm} \emph{(book ``x'' written in language ``y'')}\\
\textbf{Seen paths:}\\
/book/written\_work/previous\_in\_series \pathSeparator /book/written\_work/author \pathSeparator /people/person/nationality \pathSeparator /people/person/nationality$^{-1}$\\ \pathSeparator /people/person/languages\\
/book/written\_work/author \pathSeparator /people/ethnicity/people$^{-1}$ \pathSeparator /people/ethnicity/languages\_spoken\\
\textbf{Unseen paths:}\\
 ''in''$^{-1}$ - ''writer''$^{-1}$ \pathSeparator /people/person/nationality$^{-1}$ \pathSeparator /people/person/languages\\
/book/written\_work/author \pathSeparator addresses \pathSeparator /people/person/nationality$^{-1}$ \pathSeparator /people/person/languages\\
\hline
\textbf{Relation:}  \hspace{.2cm} /people/person/place\_of\_birth/ \hspace{.2cm}  \emph{(person ``x'' born in place ``y'')} \\
\textbf{Seen paths:}\\
 ``was,born,in'' \pathSeparator /location/mailing\_address/citytown$^{-1}$ \pathSeparator /location/mailing\_address/state\_province\_region\\
 ``from'' \pathSeparator /location/location/contains$^{-1}$\\
\textbf{Unseen paths:}\\
``born,in'' \pathSeparator /location/location/contains \pathSeparator ``near''$^{-1}$\\
``was,born,in'' \pathSeparator commonly,known,as$^{-1}$\\
\hline
\textbf{Relation:}  \hspace{.2cm} /geography/river/cities/ \hspace{.2cm}  \emph{(river ``x'' flows through or borders ``y'')} \\
\textbf{Seen paths:}\\
``at'' \pathSeparator /location/location/contains$^{-1}$\\
``meets,the'' \pathSeparator /transportation/bridge/body\_of\_water\_spanned$^{-1}$ \pathSeparator /location/location/contains$^{-1}$ \pathSeparator ``in'' \\
\textbf{Unseen paths:}\\
/geography/lake/outflow$^{-1}$ \pathSeparator /location/location/contains$^{-1}$\\
 /geography/lake/outflow$^{-1}$ \pathSeparator /location/location/contains$^{-1}$ \pathSeparator ``near''\\
\hline
\textbf{Relation:}  \hspace{.2cm} /people/family/members/ \hspace{.2cm}  \emph{(person ``y'' part of family ``x'')} \\
\textbf{Seen paths:}\\
/royalty/monarch/royal\_line$^{-1}$ \pathSeparator /people/person/children \pathSeparator /royalty/monarch/royal\_line\\ \pathSeparator /royalty/royal\_line/monarchs\_from\_this\_line\\
/royalty/royal\_line/monarchs\_from\_this\_line \pathSeparator /people/person/parents$^{-1}$ \pathSeparator /people/person/parents$^{-1}$ \pathSeparator /people/person/parents$^{-1}$\\
\textbf{Unseen paths:}\\
/royalty/monarch/royal\_line$^{-1}$ \pathSeparator ``leader''$^{-1}$ \pathSeparator ``king'' \pathSeparator ``was,married,to''$^{-1}$\\
``of,the''$^{-1}$ \pathSeparator ``but,also,of'' \pathSeparator ``married'' \pathSeparator ``defended''$^{-1}$\\
\hline
\end{tabular}
\caption{Predictive paths, according to the \emph{RNN} model, for $4$ target relations. Two examples of seen and unseen paths are shown for each target relation. Inverse relations are marked by $^{-1}$, i.e, $r(x,y) \implies r^{-1}(y,x), \forall (x,y) \in r$. Relations within quotes are OpenIE (textual) relation types.}
\label{table:paths}
}
\end{table*}
\subsection{Results}
Using our dataset, we compare the performance of the following methods:
\\
{\bf \emph{PRA Classifier}} is the method in \newcite{pra_second} which trains a logistic regression classifier by creating a feature for every path type.
\\
{\bf \emph{Cluster PRA Classifier} } is the method in \newcite{pra_recent} which replaces relation types from ClueWeb triples  with their cluster membership in the KB graph before the path finding step. After this step, their method proceeds in exactly the same manner as \newcite{pra_second} training a logistic regression classifier by creating a feature for every path type. We use pre-trained relation vectors from \newcite{limin} and use k-means clustering to cluster the relation types to $25$ clusters as done in \newcite{pra_recent}.
\\
{\bf \emph{Composition-Add}} uses a simple element-wise addition followed by sigmoid non-linearity as the composition function similar to  \newcite{bishan}.
\\
{\bf \emph{RNN-random}} is the supervised RNN model described in section \ref{sec:rnn} with the relation vectors initialized randomly.
\\
{\bf \emph{RNN}} is the supervised RNN model described in section \ref{sec:rnn} with the relation vectors  initialized using the method in \newcite{limin}.
\\
{\bf \emph{PRA Classifier-b}} is our simple extension to  the method in \newcite{pra_second}  which additionally uses bigrams in the path as  features. We add a special \emph{start} and \emph{stop} symbol to the path before computing the bigram features.
\\
{\bf \emph{Cluster PRA Classifier-b}} is  our simple extension to the  method in \newcite{pra_recent} which additionally uses bigram features computed as previously described.
\\
{\bf \emph{RNN + PRA Classifier}} combines the predictions of \emph{RNN} and \emph{PRA Classifier}. We combine the predictions by assigning the score of a fact as the sum of their rank in the two models after sorting them in ascending order.
\\
{\bf \emph{RNN + PRA Classifier-b}}  combines the predictions of \emph{RNN} and \emph{PRA Classifier-b} using the technique described previously.

Table \ref{table:results} shows the results of our experiments. The method described in \newcite{vector_pra} is not included in the table since the publicly available implementation does not scale  to our large dataset.
First, we show that it is better to train the models using all the path types instead of  using only the top $1,000$  path types as done in previous work \cite{pra_recent,vector_pra}. We can see that the RNN model performs significantly better than the baseline methods of \newcite{pra_second} and  \newcite{pra_recent}. The performance of the RNN model is not affected by initialization since using random vectors and pre-trained vectors results in similar performance.

 A surprising result is the impressive performance of our simple extension to the classifier approach. After the addition of bigram features, the naive PRA method is as effective as the Cluster PRA method. The small difference in performance between \emph{RNN} and both \emph{PRA Classifier-b} and \emph{Cluster PRA Classifier-b} is not statistically significant. We conjecture that our method has substantially different strengths than the new baseline.  While the classifier with bigram features has an ability to accurately memorize important  local structure, the RNN model generalizes better to unseen paths that are very different from the paths seen is training.   Empirically, combining  the predictions of \emph{RNN} and \emph{PRA Classifier-b} achieves a statistically significant gain over  \emph{PRA Classifier-b}.
\begin{table*}[t!]
\centering
{
\begin{tabular}{ |l|| p{3cm}|| p{3cm}| }
\hline
 & \multirow{2}{*}{\parbox[t]{3cm}{train with \\ top 1000 paths}} & \multirow{2}{*}{\parbox[t]{3cm}{train with \\ all paths}} \\  
& & \\  \hline \hline
Method & MAP & MAP  \\ \hline
\emph{PRA Classifier} & 43.46 & 51.31   \\ 
\emph{Cluster PRA Classifier}  & 46.26 & 53.23   \\ \hline \hline
\emph{Composition-Add}  & 40.23 & 45.37 \\ \hline \hline
\emph{RNN-random} & 45.52 & 56.91 \\ 
\emph{RNN} & 46.61 & 56.95  \\ \hline \hline
\emph{PRA Classifier-b} & 48.09 & 58.13  \\ 
\emph{Cluster PRA Classifier-b}  & 48.72 & 58.02   \\ \hline \hline
\emph{RNN + PRA Classifier} & 49.92 & 58.42  \\ 
\emph{RNN + PRA Classifier-b} & 51.94 & {\bf 61.17} \\ \hline
\end{tabular}
\caption{ Results comparing different methods on $46$ types. All the methods perform better when trained using all the paths than training using the top $1,000$ paths. When training with all the paths, \emph{RNN} performs significantly ($p < 0.005$) better than  \emph{PRA Classifier} and \emph{Cluster PRA Classifier}. The small difference in performance between \emph{RNN} and both \emph{PRA Classifier-b} and \emph{Cluster PRA Classifier-b} is not statistically significant.  The best results are obtained by combining the predictions of \emph{RNN} with  \emph{PRA Classifier-b} which performs significantly ($p < 10^{-5}$) better than both \emph{PRA Classifier-b} and \emph{Cluster PRA Classifier-b}. }
\label{table:results}
}
\end{table*}
\subsubsection{Zero-shot}
Table \ref{table:zeroshot} shows the results of the zero-shot model described in section \ref{sec:zeroshot} compared with the fully supervised RNN model (section \ref{sec:rnn}) and a baseline that produces a random ordering of the test facts. We evaluate on randomly selected $10$ (out of $46$) relation types, hence for the fully supervised version we train $10$ RNNs, one for each relation type. For evaluating the  zero-shot model, we randomly split the relations into two sets of equal size and train a zero-shot model on one set and test on the other set. So, in this case we have two RNNs making predictions on relation types that they have never seen during training. As expected, the fully supervised RNN outperforms the zero-shot model by a large margin but the zero-shot model without using any direct supervision clearly performs much better than a random baseline. 
\begin{table}[t!]
{
\centering
\begin{tabular}{ |l|| p{2.5cm}|| p{2.5cm}| }
\hline
 & \multirow{2}{*}{\parbox[t]{2.5cm}{train with \\ top 1000 paths}} & \multirow{2}{*}{\parbox[t]{2.5cm}{train with \\ all paths}} \\  
& & \\  \hline \hline
Method & MAP & MAP  \\ \hline
RNN  & 43.82 & 50.10  \\ \hline 
zero-shot  & 19.28 & 20.61   \\ \hline 
Random   & \multicolumn{1}{r}{7.59} &    \\ \hline 
\end{tabular}
\caption{ Results comparing the zero-shot model with supervised RNN and a random baseline on $10$ types. RNN is the fully supervised model described in section \ref{sec:rnn} while zero-shot is the model described  in section \ref{sec:zeroshot}.  The zero-shot model without explicitly training for the target relation types achieves impressive results by performing significantly ($p < 0.05$) better than a random baseline.}
\label{table:zeroshot}
}
\end{table}
\subsubsection{Discussion}
To investigate whether the performance of the RNNs were affected by multiple local optima issues, we combined the predictions of five different RNNs trained using all the paths.  Apart from \emph{RNN} and \emph{RNN-random}, we trained three more RNNs with different random initialization and the performance of the three RNNs individually are $57.09$, $57.11$ and $56.91$.  The performance of the  ensemble is $59.16$ and their performance stopped improving after using three RNNs.
So, this indicates that even though multiple local optima affects the performance, it is likely not the only issue since the performance of the ensemble is still less than the performance of  \emph{RNN + PRA Classifier-b}.

We suspect the RNN model does not capture some of the important local structure as well as  the  classifier using bigram features. To overcome this drawback, in future work, we plan to explore compositional models that have a longer memory \cite{lstm,gated,mikolov}. We also plan to include vector representations for the entities and develop models that address the issue of polysemy in verb phrases \cite{multi-rnn}.

\section{Related Work}
{\bf KB Completion} includes methods such as \newcite{dirt}, \newcite{resolver} and \newcite{entailment} that learn inference rules of length one.  \newcite{horn_clauses}  learn general inference rules by considering the set of all paths in the KB and selecting paths that satisfy a certain precision threshold. Their method does not scale well to modern KBs and also depends on carefully tuned thresholds. \newcite{pra} train a simple logistic regression classifier with NELL KB paths as features to perform KB completion while \newcite{pra_recent} and \newcite{vector_pra} extend it by using pre-trained relation vectors to overcome feature sparsity. Recently, \newcite{bishan} learn inference rules using simple element-wise addition or multiplication as the composition function.
\\
{\bf Compositional Vector Space Models} have been developed to represent phrases and sentences in natural language as vectors  \cite{lapata,baroni,ainur}.  Neural networks have been successfully used to learn vector representations of phrases using the vector representations of the words in that phrase. Recurrent neural networks have been used for many tasks such as language modeling \cite{rnnlm}, machine translation \cite{rnnmt} and parsing \cite{rnnparse}. Recursive neural networks, a more general version of the recurrent neural networks  have been used for many tasks like parsing \cite{parsing}, sentiment classification \cite{semanticRNN,sentimentRNN,irsoy}, question answering \cite{qa} and natural language logical semantics \cite{inference}.  Our overall approach is similar to RNNs with attention \cite{attentionmt,attentiongraves} since we select a  path among the set of paths connecting the entity pair to make the final prediction.
\\
{\bf Zero-shot or zero-data  learning} was introduced in \newcite{zeroshot} for character recognition and drug discovery. \newcite{zero} perform zero-shot learning for neural decoding while there has been plenty of work in this direction for image recognition \cite{zerosocher,devise,samy}.

\section{Conclusion}
We develop a compositional vector space model for knowledge base completion using recurrent neural networks. 
In our challenging large-scale dataset available at \url{http://iesl.cs.umass.edu/downloads/inferencerules/release.tar.gz}, 
our method outperforms two baseline methods  and performs competitively with a modified stronger baseline. The best results are obtained by combining the predictions of our model with the predictions of the modified baseline which achieves a $15$\%  improvement over \newcite{pra_recent}. We also show that our model has the ability to perform zero-shot inference.
\section*{Acknowledgments}
We thank Matt Gardner for releasing the PRA code, and for answering numerous question about the code and data. We also thanks the Stanford NLP group for releasing the neural networks code.   This work was supported in part by the Center for Intelligent Information Retrieval, in part by DARPA under agreement number FA8750-13-2-0020, in part by an award from Google, and in part by NSF grant \#CNS-0958392. The U.S. Government is authorized to reproduce and distribute reprints for Governmental purposes notwithstanding any copyright notation thereon. Any opinions, findings and conclusions or recommendations expressed in this material are those of the authors and do not necessarily reflect those of the sponsor.

\bibliographystyle{acl}
\bibliography{acl2014}

\end{document}